\documentclass[letterpaper]{article} % DO NOT CHANGE THIS
\usepackage{aaai24}  % DO NOT CHANGE THIS
\usepackage{times}  % DO NOT CHANGE THIS
\usepackage{helvet}  % DO NOT CHANGE THIS
\usepackage{courier}  % DO NOT CHANGE THIS
\usepackage[hyphens]{url}  % DO NOT CHANGE THIS
\usepackage{graphicx} % DO NOT CHANGE THIS
\usepackage{natbib}  % DO NOT CHANGE THIS AND DO NOT ADD ANY OPTIONS TO IT
\usepackage{caption} % DO NOT CHANGE THIS AND DO NOT ADD ANY OPTIONS TO IT
\usepackage{algorithm}
\usepackage{algorithmic}

\usepackage{newfloat}
\usepackage{listings}
\DeclareCaptionStyle{ruled}{labelfont=normalfont,labelsep=colon,strut=off} % DO NOT CHANGE THIS
\floatstyle{ruled}
\newfloat{listing}{tb}{lst}{}
\floatname{listing}{Listing}
\usepackage{multirow}
\usepackage{booktabs}
\usepackage{amsmath}
\usepackage {pifont}
\usepackage{graphicx}
\usepackage{subfigure} 
\usepackage{bm}
\nocopyright
\title{Hybrid-Supervised Dual-Search: Leveraging Automatic Learning for Loss-free Multi-Exposure Image Fusion }
\author {
    Guanyao Wu\textsuperscript{\rm 1},
    Hongming Fu\textsuperscript{\rm 1},
    Jinyuan Liu\textsuperscript{\rm 2},
    Long Ma\textsuperscript{\rm 1},
    Xin Fan\textsuperscript{\rm 3},
    Risheng Liu\textsuperscript{\rm 3}
}
\affiliations {
    \textsuperscript{\rm 1}School of Software Technology, Dalian University of Technology\\
    \textsuperscript{\rm 2}School of Mechanical Engineering, Dalian University of Technology\\
    \textsuperscript{\rm 3}DUT-RU International School of Information Science \& Engineering, Dalian University of Technology\\
    rollingplainko@gmail.com, hm.fu@hotmail.com, atlantis918@hotmail.com, \\
    malone94319@gmail.com, xin.fan@dlut.edu.cn, rsliu@dlut.edu.cn
}
\usepackage{bibentry}
\begin{document}

\maketitle

\begin{abstract}
Multi-exposure image fusion (MEF) has emerged as a prominent solution to address the limitations of digital imaging in representing varied exposure levels. Despite its advancements, the field grapples with challenges, notably the reliance on manual designs for network structures and loss functions, and the constraints of utilizing simulated reference images as ground truths. Consequently, current methodologies often suffer from color distortions and exposure artifacts, further complicating the quest for authentic image representation. In addressing these challenges, this paper presents a Hybrid-Supervised Dual-Search approach for MEF, dubbed \textbf{HSDS-MEF}, which introduces a bi-level optimization search scheme for automatic design of both network structures and loss functions. More specifically, we harnesses a unique dual research mechanism rooted in a novel weighted structure refinement architecture search. Besides, a hybrid supervised contrast constraint seamlessly guides and integrates with searching process, facilitating a more adaptive and comprehensive search for optimal loss functions. We realize the state-of-the-art performance in comparison to various competitive schemes, yielding a 10.61\% and 4.38\% improvement in Visual Information Fidelity (VIF)
for general and no-reference scenarios, respectively, while providing results with high contrast, rich details and colors.

%Leveraging a pruning-inspired neural search structure unit, our method integrates a mixed-supervised contrast constraint to guide the loss function search. Through this approach, we aim to optimize computational resources, enhance the model's generalization capabilities, and improve fusion quality across varied exposure scenarios.
\end{abstract}

\section{Introduction}

\begin{figure}[t]
	\centering
	\includegraphics[width=0.48\textwidth]{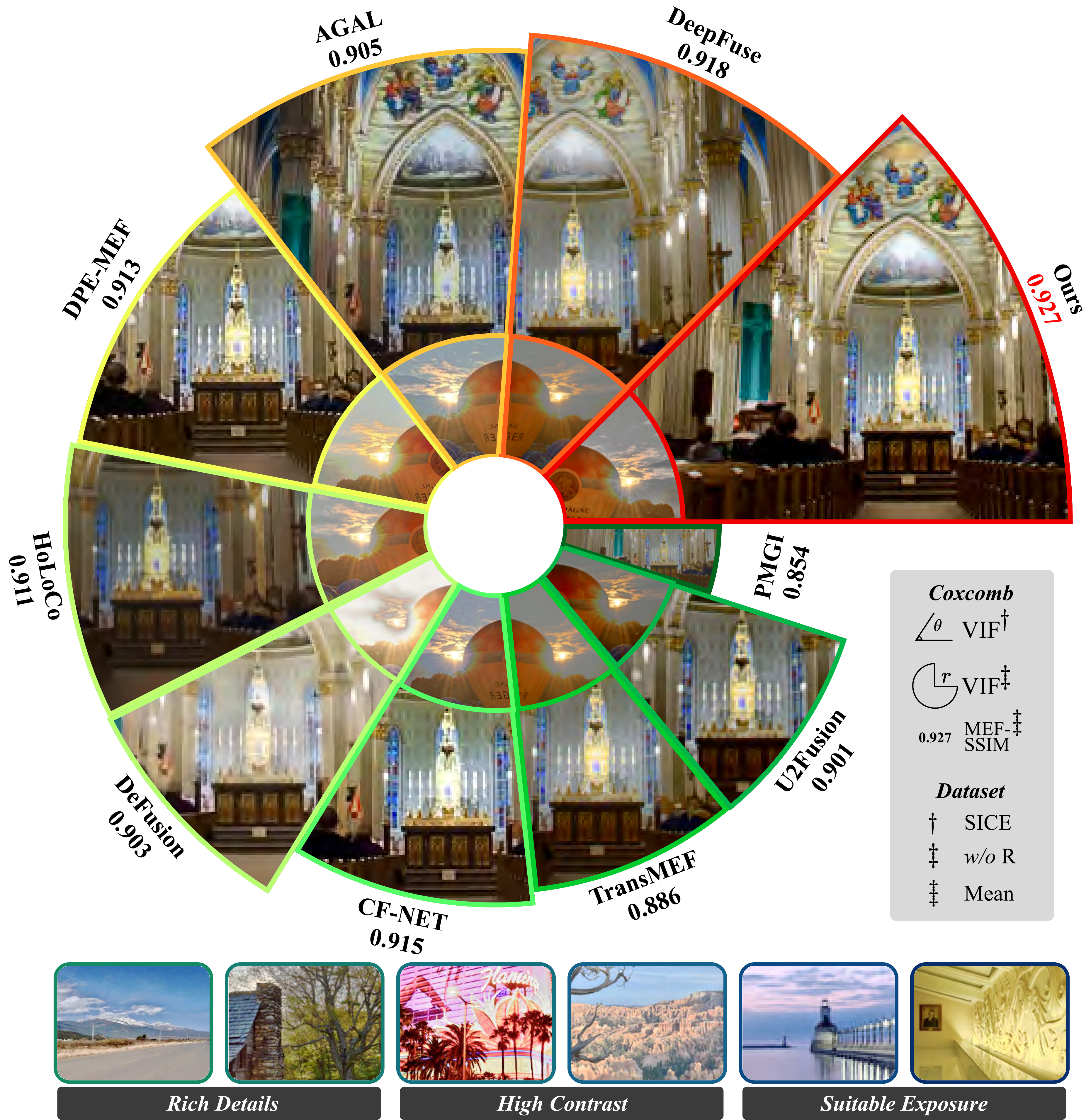} % Reduce the figure size so that it is slightly narrower than the column.
	\caption{The coxcomb chart of visual results  and key evaluation metrics. Our approach demonstrates the most visually pleasing and well-exposed fused results, with rich details that closely resemble human perception.}
	\label{fig:first}
\end{figure}

High dynamic range imaging (HDRI) has garnered extensive attention and research in recent years, aiming to accurately represent scenes from the direct sunlight to the darkest shadows found in the real world. The sensors in contemporary digital imaging devices capture a color spectrum far narrower than the intrinsic dynamic range of natural scenes~\cite{shen2012qoe}. This discrepancy leads to low dynamic range (LDR) images that often fall short in aspects of luminance and contrast, especially suffering from detrimental effects under extreme over-/under-exposure scenarios.

Rather than resorting to the costly and less efficient solution of developing high dynamic range imaging hardware to overcome traditional sensor limitations, e.g., ~\cite{nayar2000high}, multi-exposure image fusion (MEF) has emerged as a primary alternative for HDRI. This approach, propelled by advancements in digital image processing, showcases extensive applicability across diverse domains, such as image enhancement~\cite{zhao2021retinexdip, liu2020real, jiang2022target, ma2021learning, ma2022toward, ma2023bilevel}, remote sensing~\cite{palsson2017multispectral, wu2019essential}, super-resolution images~\cite{zhu2021lightweight}, object detection~\cite{wang2021cgfnet, piao2019depth, piao2020a2dele, zhang2020select}, etc. By merging a series of LDR images with varying exposure levels, it sidesteps hardware constraints, producing images more congruent with human visual perception. Nevertheless, in spite of the many methodologies that have surfaced driven by deep learning, there still remain two significant challenges in the current techniques.

%The existing sensors of digital imaging devices capture a color range much smaller than the actual dynamic range of natural scenes, which results in unsatisfactory luminance, contrast, and other imaging aspects, suffering significant damage under over-/under-exposure. With the advancement of digital image processing techniques, multi-exposure image fusion~(MEF) has become a primary solution to this issue, exhibiting a broad application prospect across various fields. By fusing images with different exposure levels, we can obtain visuals more aligned with human visual perception. However, despite the emergence of many methods driven by deep learning in this field, two major problems persist in the current technologies.

First, existing multi-exposure image fusion methods predominantly rely on manually designed network structures and loss functions, such as DeepFuse~\cite{ram2017deepfuse}, PMGI~\cite{Zhang2020RethinkingTI}, TransMEF~\cite{qu2022transmef}, etc. On one hand, such manually crafted network structures are often limited by the prior experience and knowledge, which lack flexibility, struggle to adapt to the diverse requirements of blending various exposure levels, and can produce artifacts or color distortions. Even more crucially, these designs may not be optimal, leading to wasted computational resources and decreased efficiency. On another critical note, the loss function, serving as the ultimate optimization objective for network learnable parameters, profoundly influences performance. Many studies in multi-exposure image fusion combine various loss functions based on MSE loss, meticulously adjust design weight coefficients for balance. These approaches not only demand substantial prior knowledge but also are time-consuming and labor-intensive. 

Due to manual intervention in these methods, model adaptability is limited. Manually designed networks~\cite{liu2021learning} may excel on specific datasets but falter under varying conditions. Additionally, as technology advances, new fusion demands and challenges (e.g. object detection and semantic segmentation) may arise, and manually designed approaches might struggle to adapt swiftly. Consequently, continuous manual adjustments are required, increasing research complexity.

Secondly, in current MEF datasets, such as the SICE dataset~\cite{cai2018learning}, the reference images are typically generated through existing algorithms and manually selected, serving as pseudo ground truths. Such selections fall short of accurately simulating real-world conditions and inadvertently cap the potential performance of supervised fusion techniques. Meanwhile, unsupervised methods tend to derive fusion rules based on features or pixel-level details of the source images. This often results in the produced fused images closely mirroring their source counterparts, limiting their efficacy in extreme exposure settings and causing artifacts and distortion.

To address these challenges, we propose a bi-level optimization search scheme that automatically designs loss functions and network structures for multi-exposure image fusion. Specifically, we construct a weighted structure refinement architecture search unit, supplemented by a structural search candidate space derived from simple sub-operations to adaptively fit the network structure. We then introduce a hybrid-supervised contrast constraint as the guiding motif for the search loss function, enabling it to automatically choose suitable components in a broad loss function search space to enhance generalization capability to various exposure. Here, the loss function search can be modeled as a choice of hyper-parameters. In short, our main contributions can be summarized as:

\begin{itemize}
    \item We introduce a bi-level optimization formulation for adaptively searching loss functions and network structures, which accurately models and describes the relationships among losses, structures and parameters. To our best knowledge, this is the first attempt of loss function searching in MEF.
    
    \item The adaptation issues caused by pseudo ground truths are explored from the perspective of guiding constraints in loss function search: the guidance of hybrid-supervised contrast constraints enables the searched components to adaptively handle images of different exposure levels, achieving high contrast while retaining rich details.
    
    \item A novel weighted structure refinement architecture search (WSRAS) is proposed to preserve the combinations of operations. This continuous structure search design takes into account the implicit associations between various operations, effectively enhancing the representational power of the model.
\end{itemize}

\section{Related Works}

\subsection{Deep Learning based MEF Methods}
Deep learning~\cite{liu2012fixed, ma2022practical, liu2022target, liu2023multi, jiang2022towards,liu2021searching,liu2023paif,liu2023bilevel} has shown remarkable progress in the realm of multi-exposure image fusion (MEF) due to its robust feature extraction capabilities hinged on neural networks. In recent years, many MEF methods based on deep learning have emerged.  DeepFuse~\cite{ram2017deepfuse} introduced deep learning into the MEF task for the first time. MEF-GAN~\cite{xu2020mef} applied generative adversarial networks and self-attention mechanism. Then, transMEF~\cite{qu2022transmef} utilized the transformer block and self-supervised multi-task learning, while MEF-CL~\cite{xu2023unsupervised} applied the contrastive learning, etc. But their network structure and loss function are completely dependent on manual design. This is not only time-consuming and laborious, but also requires the rich experience and knowledge of researchers. Moreover, it is impossible to accurately find the best network structure and make the network converge to the best state, which will also cause the performance of the network to decline. Although U2Fusion~\cite{xu2020u2fusion} has proposed a preliminary study on loss function balance, it is still not capable of fine adjustment of network structure and loss function weight, nor can it avoid a series of problems caused by manual design. 

\subsection{Network Architecture \& Loss Function Search}

With the rise of Neural Architecture Search (NAS), a series of methods have been developed, reinforcement learning~\cite{guo2019irlas}, evolutionary algorithms~\cite{chen2019renas}, and differentiable search methods.
Among them, differentiable search strategy~\cite{liu2018darts, liu2021investigating, liu2021smoa} has received high attention and application for various image enhancement tasks, such as  low-light enhancement~\cite{liu2021retinex},  image restoration~\cite{zhang2020memory}, and multi-exporsure fusion, etc. However, a common limitation in current methods is that they focus only on the impact of the architecture of the search on the model, ignoring the interactions and implicit associations of the minor structure of the search process. Thus, we propose a search method aimed at weakening redundant connections and grasping internal relationships, as well better preserving the optimal result of the model search.

On the other hand, the loss function also plays a key role in the model, so it has also attracted wide attention.
It is worth noting that a few methods~\cite{tang2021autopedestrian} suggest the use of automation techniques to achieve the task of dynamically and adaptively selecting the appropriate loss function, while others~\cite{wang2020loss} give a unified formula for the loss of variant softmax and search for the best loss using a reward-guided strategy. Additionally,  AutoLoss-Zero~\cite{li2020auto} proposes to use variation evolution algorithm to find loss function. Different from the above methods, we applied the hybrid contrast constraint as the evaluation criterion for the search of loss, taking into account the characteristics of MEF task.

%Drawing inspiration from the influential NAS, various numerical approaches [29]–[31] have emerged, focusing on searching for efficient model architectures tailored for multi-exposure image fusion (MEF). Specifically, [26] transcends the discrete search space by morphing it into a continuous landscape using architecture parameters. Delving deeper into unsupervised territories, [29] integrates unsupervised techniques into NAS, challenging the conventional reliance on labels. The groundbreaking work by [30] showcases a convergence-centric NAS mechanism, uniquely driven by randomized labels during the search process. A few pioneering efforts have also ventured into harnessing the potential of NAS explicitly for MEF. For instance, [34] formulates a state-of-the-art framework to autonomously refine attention mechanisms tailored for MEF. Yet, a recurring limitation across these methodologies is their singular focus on the architectural impact for MEF, sidelining other pivotal factors.
%
%On the other hand, loss functions also have significant impact on deep models, thus have stimulated the interest of researchers recently. Noticeably, a few methods [15]–[19], [35] propose to utilize automated techniques towards dynamically
%and adaptively selecting appropriate loss functions for various tasks. [18] develops a unified formulation for variant softmax losses and leverages a reward-guided strategy for searching the optimal loss for face detection. AutoLoss-Zero [16] proposes to discover loss functions via a variant evolutionary algorithm for image semantic segmentation.

\section{The proposed Method}

\subsection{Problem Formulation}
We initially lay out the main components: image fusion network $\mathcal{F}$ and the source images as $\mathbf{I}_{\mathtt{o}}$, $\mathbf{I}_{\mathtt{u}}$, and  fused image as $\mathbf{I}_{\mathtt{f}}$. The fusion process can be symbolized as $\mathbf{I}_{\mathtt{f}}=\mathcal{F}(\mathbf{I}_{\mathtt{o}}, \mathbf{I}_{\mathtt{u}}; \bm{\omega}; \bm{\alpha})$. The fusion loss $\mathcal{L}^{\mathtt{f}}$ is introduced to facilitate the training of fusion network as $\mathcal{L}^{\mathtt{f}}\big(\mathcal{F}(\mathbf{I}_{\mathtt{o}}, \mathbf{I}_{\mathtt{u}}; \bm{\omega}; \bm{\alpha});\bm{\beta})\big)$,  where $\bm{\omega}_{\mathtt{f}}$ denotes the trainable parameters of $\mathcal{F}$ and $\bm{\alpha}$/$\bm{\beta}$ represent the desired architecture/loss parameters determined in the defined search space $\mathcal{A}$/$\mathcal{B}$. 

Overall, the specific nested form of the bi-level optimization for MEF is formulated as follows: 

\begin{eqnarray}
	\begin{split}
		\quad\min\limits_{\bm{\alpha}, \bm{\beta}} \mathcal{L}_{\mathtt{search}}\big(\bm{\beta}; \mathcal{F}(\mathbf{I}_{\mathtt{o}}, \mathbf{I}_{\mathtt{u}}; \bm{\omega}_{\mathtt{f}}^*; \bm{\alpha})\big),\\
		s.t. \ \bm{\omega}^* \in \arg\min\limits_{\bm{\omega}} \mathcal{L}^{\mathtt{f}}_{\mathtt{train}}\big( \mathcal{F}(\mathbf{I}_{\mathtt{o}}, \mathbf{I}_{\mathtt{u}}; \bm{\omega}; \bm{\alpha});\bm{\beta}\big).
	\end{split}
	\label{eq:search}
\end{eqnarray}
The upper-level objective is to minimize the weighted parameters with design requirements based on the measurements of searching loss. The lower-level optimization is to acquire the desired fusion parameters based on the given $\bm{\alpha}$ and guided by the loss function composed of $\bm{\beta}$.

At its core, the bi-level optimization intuitively captures the intrinsic relationship and dependencies between the network structure and its corresponding loss function. Such a modeling strategy leverages the mutual feedback between these two levels to derive more effective and robust solutions. In contrast to traditional optimization methods, this bi-level approach allows for a simultaneous search of both architecture and loss function parameters, ensuring that each iteration mutually informs and refines the other. This synergy ensures that the final model achieves a balanced interplay between the desired architecture and the associated loss function, leading to superior performance and adaptability in MEF task.

\subsection{Dual Searching Solution Scheme}
 From the above modeling, it is evident that the choice of $\mathcal{L}_{\mathtt{search}}$ plays a crucial role in the entire optimization process, thus requiring careful design and selection. In the process of architecture search, there exists a close relationship between the $\bm{\alpha}$ and $\bm{\omega}$. In previous research, researchers often employed the same task-specific loss ($\mathcal{L}^{\mathtt{f}}$ here) to guide the optimization of $\bm{\alpha}$, and we followed the similar approach in our design. Considering that the loss function, as the global learning objective of the network, holds a higher level of significance in the structure, we decide the hybrid-supervised contrastive constraint $\boldsymbol{\Gamma}_{\mathtt{H}}$ (will be introduced below) as the validation to evaluate the effectiveness of the obtained loss function during the search process. The specific solution scheme is represented as follows:

\begin{equation}
	\left\{\begin{array}{l}
		\boldsymbol{\beta}^{t+1}=\min\limits_{\boldsymbol{\beta} \in \mathcal{B}} \boldsymbol{\Gamma}_{\mathtt{H}}\left(\boldsymbol{\beta}, \boldsymbol{\omega} ; \boldsymbol{\alpha}^t\right), \\
		\boldsymbol{\alpha}^{t+1}=\min\limits_{\boldsymbol{\alpha} \in \mathcal{A}} \mathcal{L}_{\mathrm{val}}^{\mathtt{f}}\left(\boldsymbol{\alpha}, \boldsymbol{\omega} ; \boldsymbol{\beta}^{t+1}\right),
	\end{array}\right.
\end{equation}
where $t$ denotes the iterations. In order to effectively solve Eq.~\eqref{eq:search}, inspired by the remarkably success of  differentiable architecture search~\cite{liu2018darts}, we adopt the first-order one-step truncated approximation strategy to compute the gradients of $\bm{\alpha}$ and $\bm{\beta}$. Also, the whole procedure is shown in Algorithm~\ref{alg:1}.

\begin{algorithm}[htb]
	\caption{Dual Search for Structure and Loss Function}\label{alg:joint_search}
	\begin{algorithmic}[1]
		\REQUIRE Fusion loss \(\mathcal{L}^{\mathtt{f}}\), hybrid-supervised contrast constraint \(\boldsymbol{\Gamma}_{\mathtt{h}}\), search spaces \(\mathcal{A}\) and \(\mathcal{B}\), and other necessary hyper-parameters.
		\ENSURE Optimal parameters \(\bm{\alpha}^{*}\) and \(\bm{\beta}^{*}\).
		
		\WHILE{not converged}
		\STATE \% Optimizing the image fusion network
		\STATE \(\bm{\omega}^\mathtt{f} \leftarrow \bm{\omega}^\mathtt{f}-\nabla \mathcal{L}^\mathtt{f}_{\mathtt{train}}(\mathbf{I}_{\mathtt{o}},\mathbf{I}_{\mathtt{u}};\bm{\omega}^\mathtt{f},\bm{\beta})\)
		
		\STATE \% First-order approximation to optimize \(\bm{\beta}\)
		\STATE \(\bm{\beta} \leftarrow \bm{\beta} - \nabla_{\bm{\beta}}\boldsymbol{\Gamma}_{\mathtt{H}}\left(\boldsymbol{\beta}, \boldsymbol{\omega} ; \boldsymbol{\alpha}\right)\)
		
		\STATE \% First-order approximation to optimize  \(\bm{\alpha}\)
		\STATE \(\bm{\alpha} \leftarrow \bm{\alpha} - \nabla_{\bm{\alpha}}\mathcal{L}_{\mathrm{val}}^{\mathtt{f}}\left(\boldsymbol{\alpha}, \boldsymbol{\omega} ; \boldsymbol{\beta}\right)\)
		
		\WHILE{\( |\mathcal{A}| > P \)}
		\IF{\(\mathrm{min} \bm{\alpha}_i < \theta \)}
		\STATE \% Prune the operation with the smallest weight in search space \( \mathcal{A} \)
		\STATE \(\mathcal{A} \leftarrow \text{RefineSmallestWeight}(\mathcal{A})\)
		\ENDIF
		\ENDWHILE
		
		\ENDWHILE
		
		\RETURN Top-$P$ operations based on \(\bm{\alpha}^{*}\), \(\bm{\beta}^{*}\).
	\end{algorithmic}
	\label{alg:1}
\end{algorithm}

\subsection{Hybrid Contrastive Constraint}
In the field of MEF, contrastive learning has demonstrated significant application potential. We choose it as the evaluation criterion to guide the search for the loss function since its unique advantage lies in the model being trained to be drawn to positive samples and repelled from negative ones. This design deviates from traditional fusion losses, enabling a higher-level and deeper extraction of discriminative features. Furthermore, we propose a hybrid supervision strategy, where the positive samples include both natural light images and reference images from the dataset. This design allows the model to draw rich information from diverse positive samples, ensuring the generation of highly natural and authentic fusion results under various environmental and lighting conditions.

In order to get the potential feature representation of images, we apply the pre-trained VGG-16~\cite{simonyan2014very} network, denoted as ${G}$.
We introduced simulated reference images and natural images, denoted as $\mathbf{I}_{\mathtt{r}}$ and $\mathbf{I}_{\mathtt{n}}$, as hybrid-supervised contrast constraints to serve as positive samples, respectively.  (i.e.$\mathbf{I}_{\mathtt{p}}$=\{$\mathbf{I}_{\mathtt{r}}$, $\mathbf{I}_{\mathtt{n}}$\}) The source images ($\mathbf{I}_{\mathtt{u}}$ \& $\mathbf{I}_{\mathtt{o}}$) are used as negative samples. We construct the following expression as the contrastive constraint:

\begin{equation}
	\begin{aligned}
		\boldsymbol{\Gamma} _{\mathtt{P}} =  \sum\limits_{i=1}^{N} \frac {
			\left\|\mathbf{F}_{\mathtt{i}}-\mathbf{P}_{\mathtt{i}}\right\|_2	}
		{\sum_{m}^{M}  (\left\|\mathbf{F}_{\mathtt{i}}-\mathbf{u}_{\mathtt{i}}^{\mathtt{m}}\right\|_2 + \left\|\mathbf{F}_{\mathtt{i}}-\mathbf{o}_{\mathtt{i}}^{\mathtt{m}}\right\|_2 )}\\
	\end{aligned}
\end{equation}

where $N$ and $M$ stands for the number of layers of VGG and negative samples, separately. $\mathbf{F}_{\mathtt{i}}$ denotes the feature of the fused image at the $i$th layer of ${G}$, which is defined as $G_{\mathtt{i}}(\mathbf{I}_{\mathtt{f}})$.  $m$ means the $m$th negative sample. Similarly, $\mathbf{P}_{\mathtt{i}}$,  $\mathbf{u}_{\mathtt{i}}^{\mathtt{m}}$,  $\mathbf{o}_{\mathtt{i}}^{\mathtt{m}}$ corresponds to $\mathbf{I}_{\mathtt{p}}$,  $\mathbf{I}_{\mathtt{u}}^{\mathtt{m}}$ and $\mathbf{I}_{\mathtt{o}}^{\mathtt{m}}$, separately.  $\left\| \;\cdot\; \right\|_2$ denotes mean square error (MSE). 
Finally, the total hybrid contrastive constraint can be defined as:
$\boldsymbol{\Gamma}_{\mathtt{H}} = \boldsymbol{\Gamma}_{\mathtt{R}} + \boldsymbol{\Gamma}_{\mathtt{N}}.$ 

\subsection{Weighted Structure Refine Architecture Search}

As alluded to earlier, our framework is built upon DARTS, in which the operation with the highest weight is directly chosen to construct the final network architecture upon convergence of the search. However, such a methodology overlook the interplay between sub-operations in the weighted sum of mixed operations and could potentially hinder the latent performance of the final structure. To delve deeper into the potential of sub-operations within mixed operations, we propose the Weighted Structure Refinement Architecture Search (WSRAS) approach. The essence of WSRAS lies in its two specific strategies: the pruning refine and weight retention operation.
\subsubsection{Pruning Refinement}
Throughout the search, we continuously eliminate sub-operations with the lowest weights. Specifically, if the weight of a sub-operation is the minimum for the current node and falls below a predetermined threshold, it is pruned. This not only avoids potential interference from redundant sub-operations in the search but also minimizes the mutual influence between sub-operations, thus enhancing search efficiency.
\subsubsection{Weight Retention}
Instead of merely selecting the operation with the highest weight at the end of our search, we opt to retain the top $P$ operations based on their weights and integrate them into the final network architecture. The benefit of this approach is twofold: it maintains a certain level of network complexity while capitalizing on the complementary nature between various operations. This method thoroughly considers the implicit connections between operations, effectively boosting the power of the model.

\subsection{Overall Workflow}
\begin{figure}[t]
	\centering
	\includegraphics[width=0.48\textwidth]{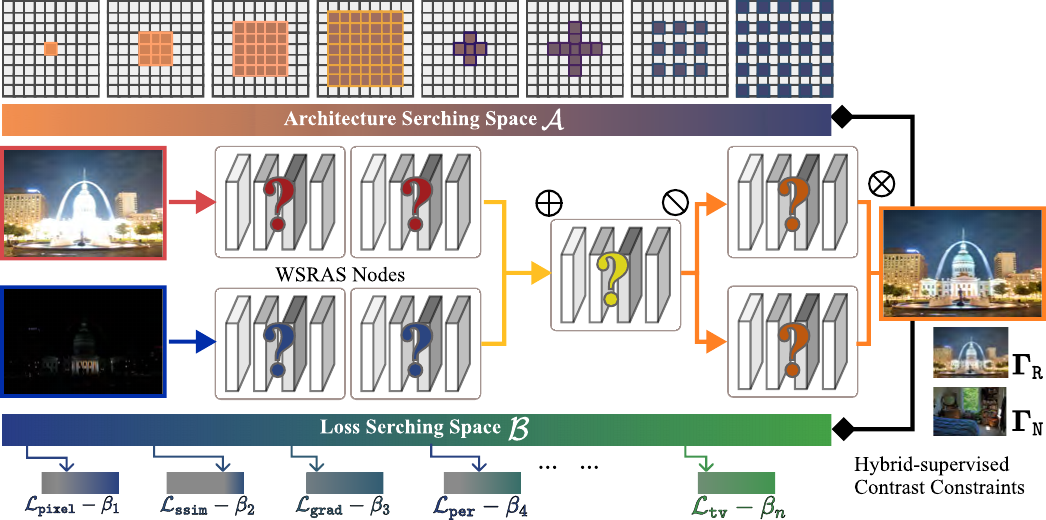} % Reduce the figure size so that it is slightly narrower than the column.
	\caption{The overall workflow.}
	\label{fig:workflow}
\end{figure}

The network consists of two main components: the Feature Attention Module and a Dual-Stream Processing Network inspired by the Retinex theory. The overall workflow is shown in Figure~\ref{fig:workflow}.
In the attention module, images with different exposures are first transformed into feature representations and subsequently integrated to form a comprehensive feature representation. Then the dual-stream processing network, grounded on the Retinex theory, decomposes the image into intensity and illumination components. The intensity component undergoes iterative enhancement, while the illumination component is refined through repetitive optimization.
%\footnote{The visualization and ablation of specific structures are attached to the supplementary material.}.
 The final results of both components are multiplied and transformed to yield the final fused image. All intermediate nodes are searchable weight-retention nodes based on WSRAS.

\begin{figure*}[t]
	\centering
	\includegraphics[width=0.98\textwidth]{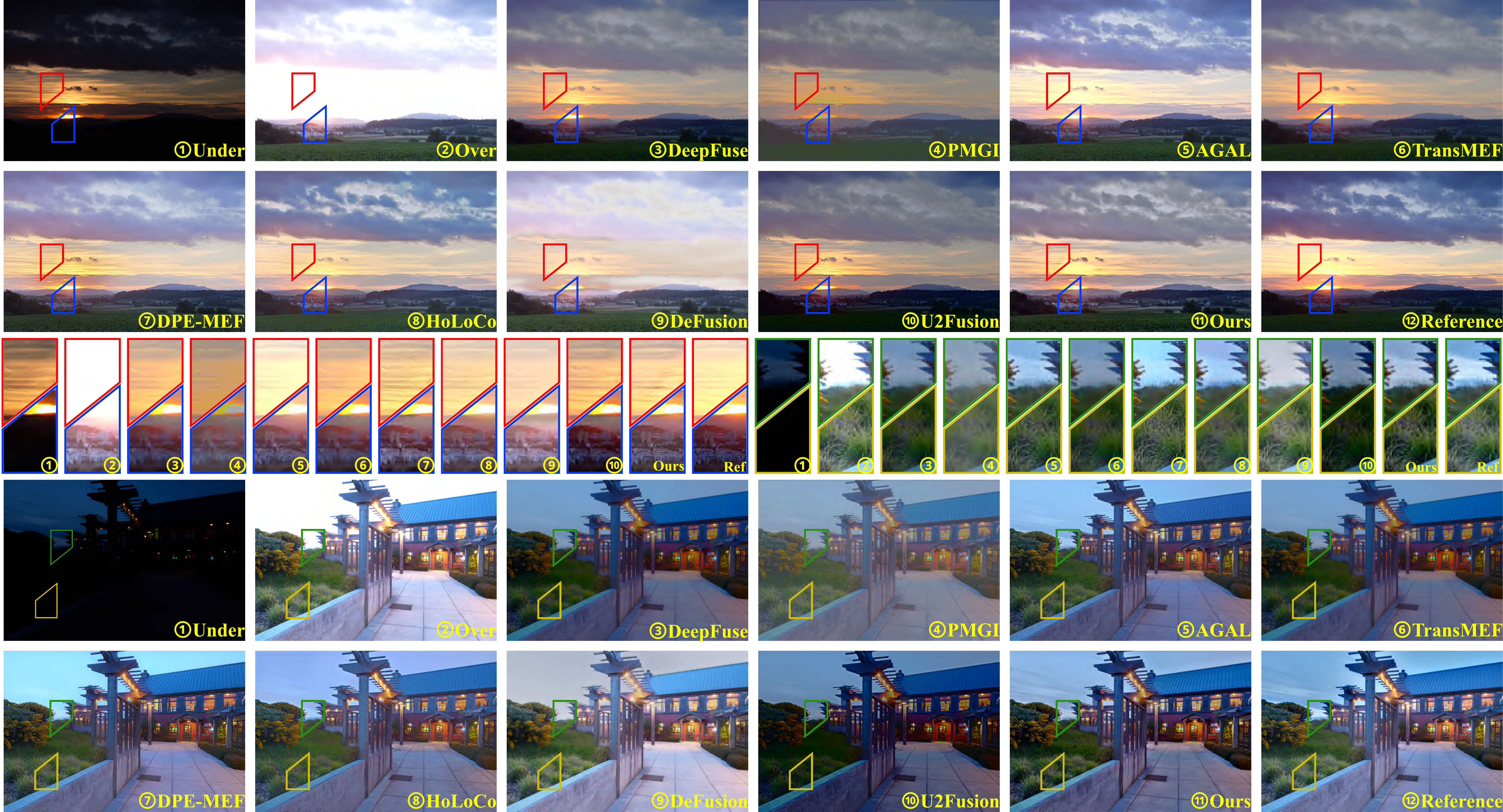} 
	\caption{Qualitative comparison with other state-of-the-art methods on the SICE dataset with reference images.}
	\label{fig:mainsice}
\end{figure*}

\subsection{Search Space of Architecture and Loss}
\subsubsection{Loss Search Space}
Our search space for training loss encompasses an image quality-oriented loss set, including pixel-level metrics like $\mathcal{L}_1$, $\mathcal{L}_{\mathtt{MSE}}$, and PSNR loss, ensuring fusion image pixel intensity consistency with source images. Structural metrics, such as SSIM~\cite{wang2004image} and MEF-SSIM~\cite{ma2019deep}, gauge brightness, contrast, and structural similarities. Gradient loss, widely recognized in works like ~\cite{sun2022detfusion, ma2021stdfusionnet}, maintains texture details from source images. Additionally, we incorporate perceptual loss~\cite{johnson2016perceptual} for feature domain disparities, color loss~\cite{wang2019underexposed} for color consistency with reference images, and TV loss~\cite{osher2005iterative} to minimize noise.  The detailed formulations are depicted in Table.~\ref{tab:lossspace2} and $\mathbf{R}$ represents the relevant images. Unlike typical loss function designs, we also adhere to the principle of hybrid supervision in the design of $\mathbf{R}$ to better align with the modeling requirements and effectively search for suitable loss function.

\subsubsection{Architecture Search Space}
In designing the network architecture, we judiciously opted for a set of lightweight operations, striving to strike a balance between enhancing search efficiency and minimizing latent correlations introduced by intricate operations. Although we did not incorporate intricate operation primitives, the inclusion of weight-preserving operations facilitates more probable coordination and combinations among operations. Specifically, our operation set comprises convolutions with kernel sizes of 1$\times$1, 3$\times$3, 5$\times$5, 7$\times$7, as well as asymmetrical kernels of 1$\times$3, 3$\times$1, 1$\times$5, and 5$\times$1. Additionally, we incorporated dilated convolutions of sizes 3$\times$3, 5$\times$5, and 7$\times$7 with a dilation rate set to 2.

\begin{table}[]
	\centering
	\small
	\renewcommand{\arraystretch}{1.4}
	\setlength{\tabcolsep}{2.4mm}{
		%		\begin{tabular}{|c|c||c|c|}
			\begin{tabular}{ll}
				\bottomrule
				Expression                              &Candidate set of $\mathbf{R}$                                                                           \\
				\toprule
				\bottomrule
				
				\ding{172} $\mathcal{L_{\mathtt{pixel}}:}\left\|\mathbf{I}_{\mathtt{f}}-\mathbf{R}\right\|_\mathtt{1/2}$ & $\mathbf{I}_\mathtt{o}$, $\mathbf{I}_\mathtt{u}$, $\mathbf{I}_{\mathtt{r}}$\\\hline
				\ding{173} $\mathcal{L_{\mathtt{ssim}}:}1 - \mathtt{SSIM}\,or\,\mathtt{MEF\,SSIM} (\mathbf{I}_{\mathtt{f}}, \mathbf{R})$ & $\mathbf{I}_\mathtt{o}$, $\mathbf{I}_\mathtt{u}$, $\mathbf{I}_{\mathtt{r}}$\\\hline
				
				\ding{174}$\mathcal{L_{\mathtt{grad}}:}\left\|\nabla \mathbf{I}_{\mathtt{f}}- \mathbf{R}\right\|_1$ & $\operatorname{max}(\nabla\mathbf{I}_{\mathtt{o}}, \nabla\mathbf{I}_{\mathtt{u}})$
				\\\hline
												%\cellcolor{orange!25}
				\ding{175}$\mathcal{L_{\mathtt{per}}:}\left\|\varphi_l(\mathbf{I}_{\mathtt{f}})-\varphi_l(\mathbf{R})\right\|_2$ &  $\mathbf{I}_\mathtt{o}$, $\mathbf{I}_\mathtt{u}$, $\mathbf{I}_{\mathtt{r}}$
				\\\hline
								
				\ding{176}$\mathcal{L_{\mathtt{psnr}}:}-\mathtt{PSNR}(\mathbf{I}_{\mathtt{f}}, \mathbf{R})$ &  $\mathbf{I}_{\mathtt{r}}$
				\\\hline
				
				\ding{177}$\mathcal{L_{\mathtt{color}}:}\sum_{i}\angle((\mathbf{I}_{\mathtt{f}})_{i}, (\mathbf{R})_{i})$ &  $\mathbf{I}_{\mathtt{r}}$
				\\\hline
				
				\ding{178}$\mathcal{L_{\mathtt{tv}}:}\frac{1}{H W}((\nabla_{x}\mathbf{I}_{\mathtt{f}})^{2} + (\nabla_{y}\mathbf{I}_{\mathtt{f}})^{2})$ & - \\
								\toprule
			\end{tabular}
		}
	\caption{Notations of the loss search space.}
	\label{tab:lossspace2}
	\end{table}

\section{Experiments}

\begin{figure*}[t]
	\centering
	\includegraphics[width=0.98\textwidth]{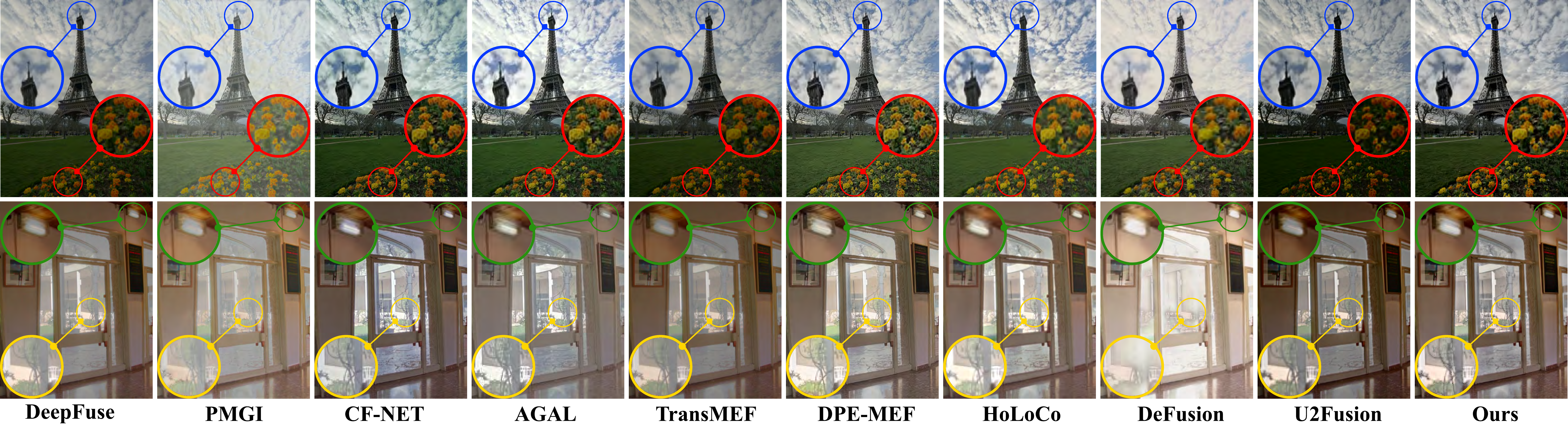}
	\caption{Qualitative comparison with other state-of-the-art methods on the  dataset without reference.}
	\label{fig:main31}
\end{figure*}

\begin{table*}[thb]
	\centering
	\small
	\renewcommand{\arraystretch}{1.2}
	\setlength{\tabcolsep}{0.6mm}{

			\begin{tabular}{|c|c|c c c c c c|c c c c c c|}
				\hline
				\multicolumn{2}{|c|}{Dataset} & \multicolumn{6}{|c|}{SICE} & \multicolumn{6}{|c|}{Dataset \emph{w/o} reference} \\
				\hline
				Method & Source & SD$\uparrow$ & VIF$\uparrow$ & CC$\uparrow$ & TMQI$\uparrow$ & MS-SSIM$\uparrow$ & MEF-SSIM$\uparrow$  & SD$\uparrow$ & VIF$\uparrow$ & EN$\uparrow$ & ${Q^{\mathrm{AB/F}}}$ $\uparrow$ & 	 MS-SSIM$\uparrow$ & MEF-SSIM$\uparrow$ \\ \hline
				
				Deepfuse & ICCV'17 & 9.968 &\underline{1.433} & 0.923 & 0.215 & 0.918 & 0.888 & 9.856 & 1.467 & 6.999 & 0.648  & 0.921 & 0.948\\
				PMGI & AAAI'20 & 10.000 & 0.968 & 0.895 & 0.207 & 0.847 & 0.828 &\underline{10.222} & 0.988 & 6.621 & 0.427 
				& 0.815 & 0.880\\
				CF-NET & TIP'21 & 9.922 & 1.241 &\underline{0.938} & 0.213 & 0.926 & 0.890 & 9.913 & 1.408 & 7.250 &\textbf{0.694} & 0.913 & 0.941 \\
				AGAL & TCSVT'22 & 9.782 & 1.405 & 0.927 & 0.215 &\underline{0.937} &\underline{0.898} & 9.836 &\underline{1.485} & 7.311 & 0.606  & 0.929 & 0.912\\
				TransMEF & AAAI'22 & 9.682 & 1.215 & 0.922 & 0.212 & 0.860 & 0.839 & 9.458 & 1.306 & 6.857 & 0.566 & 0.906 & 0.934 \\
				DPE-MEF & IF'22 & 10.187 & 1.316 & 0.928 & 0.214 & 0.921 & 0.877 & 9.992 &\underline{1.485} &\underline{7.321} &\underline{0.685} &\underline{0.945} & \underline{0.949}\\
				HoLoCo & IF'23 & 9.788 & 1.325 &\underline{0.938} & 0.215 & 0.925 & 0.890 & 9.610 & 1.397 & 7.163 & 0.633  & 0.922 & 0.933\\
				DeFusion & ECCV'22 &\underline{10.433} & 1.329 & 0.890 & 0.208 & 0.893 & 0.870 & 10.176 & 1.277 & 7.035 & 0.634 & 0.927 & 0.935\\
				U2Fusion & TPAMI'22 & 9.884 & 1.219 & 0.914 &\underline{0.216} & 0.897 & 0.874 & 9.635 & 1.291 & 6.796 & 0.556  & 0.902 & 0.928\\
				Ours & Proposed &\textbf{10.451} &\textbf{1.585} &\textbf{0.956} &\textbf{0.218} &\textbf{0.951} &\textbf{0.903} &\textbf{10.235} &\textbf{1.550} &\textbf{7.343} &\textbf{0.694} &\textbf{0.952} & \textbf{0.950}\\
				
				\hline
				
				\multicolumn{2}{|c|}{\emph{Improvement}} & \textbf{0.018$\uparrow$} & \textbf{0.152$\uparrow$} & \textbf{0.018$\uparrow$} & \textbf{0.002$\uparrow$} & \textbf{0.014$\uparrow$} & \textbf{0.005$\uparrow$} & \textbf{0.013$\uparrow$} & \textbf{0.065$\uparrow$} & \textbf{0.022$\uparrow$} & - & \textbf{0.007$\uparrow$} 
				& \textbf{0.001$\uparrow$}\\
				\hline

			\end{tabular} 

	}

\caption{Qualitative comparison with other state-of-the-art methods. \textbf{Bold: best}; \underline{underline: 2nd best}}
\label{tab:main1}
\end{table*}

\subsection{Implementation Details}
We conducted experiments on the SICE dataset~\cite{cai2018learning}, and randomly selected 452 pairs of images as the training set and 113 as the testing set, and select another 31 images without reference~\cite{cai2018learning, ma2017robust} as part of the testing set. At the same time, we use CoCoval2017 as a natural light image set. All images are randomly cropped to the size of 256$\times$256 during the search and training process, and all parameters are updated using the Adam optimizer. For searching, half of the training set is randomly selected as the verification set, the batch size and epoch are set to 2 and 10, the learning rate of the network structure weight, the loss function weight and the network parameter are set to 2e-1, 3e-2 and 2e-4, respectively. For training of 60 epoches, the batch size and the learning rate are set to 10 and 1e-4. The overall framework is implemented on Pytorch with an NVIDIA Tesla V100 GPU.
\subsection{Contrast Methods \& Evaluation Metrics}
We choose 9  state-of-the-art MEF methods as competitors, which include a decompisiton-based method: DeFusion~\cite{Liang2022ECCV}, and 8 deep learning-based method: DeepFuse~\cite{ram2017deepfuse}, PMGI~\cite{Zhang2020RethinkingTI}, CF-NET~\cite{deng2021deep}, AGAL~\cite{liu2022attention}, TransMEF~\cite{qu2022transmef} , DPE-MEF~\cite{HAN2022248}, HoLoCo~\cite{liu2023holoco} and U2Fusion~\cite{xu2020u2fusion}.

In evaluation, the 8 used metrics can be categorized into reference-based and no-reference-based indicators. Among the former, correlation coefficient (CC)\cite{Shah2011MultifocusAM} indicates the correlation between the fused and reference images, tone-mapped image quality Index (TMQI)\cite{6319406} represents the tonal mapping effect of HDRI images, multi-scale structural similarity index measure (MS-SSIM)\cite{1292216} evaluates the structure retention degree of the fused image across scales, and the task specific MEF-SSIM\cite{7120119}. On the other hand, no-reference metrics include standard deviation (SD)\cite{Yun-JiangRao_1997} showcasing contrast ratio, visual quality fidelity (VIF)\cite{Han2013ANI} gauging the degree of matching with human visual perception, entropy (EN)\cite{Roberts2008AssessmentOI} estimating the total information, and gradient-based (${Q^{\mathrm{AB/F}}}$)\cite{Xydeas2000ObjectiveIF} highlighting the detail quantity.

\begin{table*}[thb]
	\centering
	\small
	\renewcommand{\arraystretch}{1.3}
	\setlength{\tabcolsep}{2.5mm}{
		\begin{tabular}{|c|c c c c| c c c c |}
			\hline
			
			\multirow{2}{*}{Condition} & \multicolumn{4}{c|}{Loss Function Searching Ablation} & \multicolumn{4}{c|}{Neural Architecture Search Ablation} \\ 
			
			%				\cmidrule(lr){2-5} \cmidrule(lr){6-10} \cmidrule(lr){11-16}
			\cline{2-9}
			& \emph{w/o} $\boldsymbol{\Gamma}_{\mathtt{H}}$ & \emph{only} $\boldsymbol{\Gamma}_{\mathtt{R}}$ & \emph{only} $\boldsymbol{\Gamma}_{\mathtt{N}}$ &  Ours & \emph{only} WR & \emph{only} PR & DARTS  &  Ours \\\hline
			
			SD&9.671 & \underline{9.841} &9.761 & \textbf{10.451}$_{\uparrow0.610}$ &10.050 & \underline{10.056} &9.862  
			& \textbf{10.451}$_{\uparrow0.395}$  \\
			VIF&1.308 & \underline{1.392} &1.176 & \textbf{1.585}$_{\uparrow0.193}$ & \underline{1.481} &1.426 &1.429 
			& \textbf{1.585}$_{\uparrow0.104}$ \\
			CC&0.916 & \underline{0.936} &0.925 & \textbf{0.956}$_{\uparrow0.020}$ & \underline{0.943} &0.937 &0.941 
			& \textbf{0.956}$_{\uparrow0.013}$ \\
			TMQI &0.214  & \underline{0.215} &0.212 & \textbf{0.218}$_{\uparrow0.003}$ & \underline{0.216} & \underline{0.216} &0.215  & \textbf{0.218}$_{\uparrow0.002}$  \\
			MS-SSIM&0.909 & \underline{0.934} &0.905 & \textbf{0.951}$_{\uparrow0.017}$ & \underline{0.946} &0.935 &0.935 
			& \textbf{0.951}$_{\uparrow0.005}$\\
			MEF-SSIM &0.870  & \underline{0.882} &0.865 & \textbf{0.903}$_{\uparrow0.021}$ & \textbf{0.908} &0.889 &0.902 
			& \underline{0.903}$_{\downarrow0.005}$ \\
			
			\hline
			
		\end{tabular} 
	}
	\caption{ Quantitative results of ablation experiments on the {SICE} dataset. \textbf{Bold: best}; \underline{underline: 2nd best}.}
	\label{tab:ab}
\end{table*}

\subsection{Comparisons on the SICE Dataset}
\subsubsection{Qualitative Comparisons}
In Figure~\ref{fig:mainsice}, we present two sequences of typical visual results along with four zoomed comparison regions. Overall, certain methods (e.g. DeepFuse, U2Fusion, and TransMEF) exhibit noticeable underexposure, struggling to balance the impact of severely underexposed source images. On the other hand, some methods (e.g. AGAL and DeFusion) tend to retain images from well-exposed scenes with rich information, yet this still leads to exposure misalignment. 

Benefiting from the advantages brought by the dual searching, our approach efficiently preserves useful information from the source images, achieving retention of details and restoration of colors (evident in cloud layers in the red and green regions of Figure~\ref{fig:mainsice}). Additionally, driven by the hybrid-supervised contrastive constraint, our approach manages to surpass reference images in certain areas (as seen in the blue and yellow regions of Figure~\ref{fig:mainsice}). The high contrast in details and richer colors contribute to improved depth perception in the fused images.

\subsubsection{Quantitative Comparisons}
Table~\ref{tab:main1} on the left presents our superior results on the SICE datasets across all six metrics, with a notable enhancement in VIF, signifying our ability to offer fidelity akin to human vision. As illustrated in Figure~\ref{fig:mainsice}, our exposure emulates natural light more effectively than the reference. Moreover, our approach outperforms others in SD, CC, and MS-SSIM, ensuring greater contrast, information preservation, and image clarity.

\subsection{Comparisons on the Dataset \emph{w/o} Reference}
\subsubsection{Qualitative Comparisons}
To verify comprehensively, we further evaluate the methods using the image sequences of \emph{w/o} reference images. The qualitative results are shown in Figure~\ref{fig:main31}. 
Our results demonstrate high contrast (e.g. clouds in the blue region and light tubes in the green region), rich details, and vibrant colors (flower bed in the red region). More importantly, we exhibit significant advantages in handling novel exposure-compromised areas (yellow part of the scene behind the door, including tree trunks and architectural details). The hybrid-supervision contrast constraint and loss function search space empower us to handle a broader range of natural exposure images. This capability somewhat diminishes limiting effect of the reference images on algorithm performance, enabling the production of fused images that more closely align with human visual perception.

\subsubsection{Quantitative Comparisons}
The excellent results for metrics on this dataset are presented in the table on the right, as shown in Table~\ref{tab:main1}. These results exhibit a trend similar to that of the SICE dataset. Notably, we achieve an absolute leading advantage over all the comparative methods in the key metric MEF-SSIM in the no-reference dataset, which demonstrates the superiority of the proposed approach.

\subsection{Ablation Studies}
%\subsection{Ablation Studies\footnote{More ablation experiments are attached in the supplementary material.}}
\subsubsection{Study on Loss Function Searching}

\begin{figure}[t]
	\centering
	\includegraphics[width=0.48\textwidth]{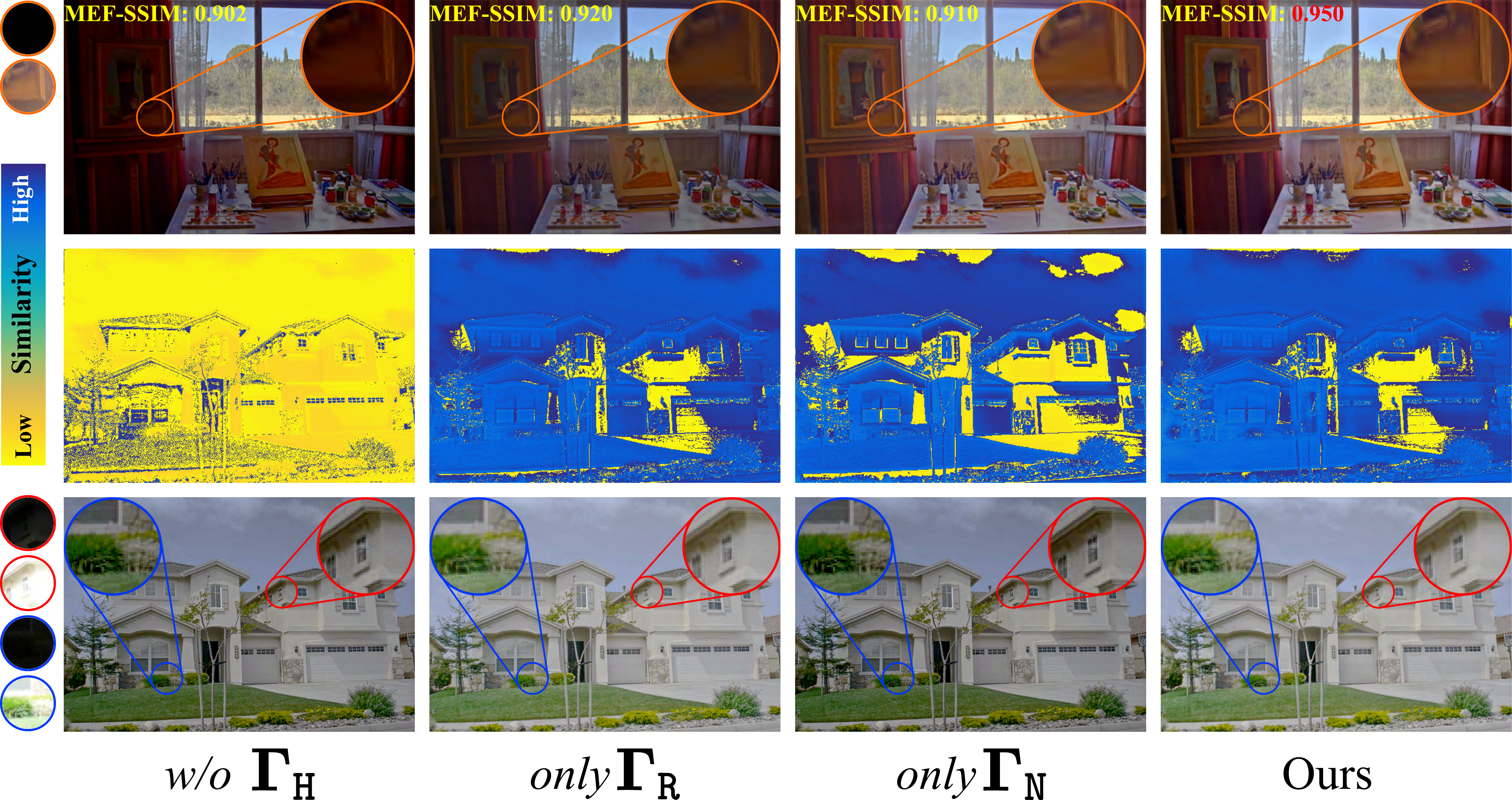} % Reduce the figure size so that it is slightly narrower than the column.
	\caption{Display of visualization results of loss function ablation variants.}
	\label{fig:Ab1}
\end{figure}

\begin{figure}[t]
	\centering
	\includegraphics[width=0.48\textwidth]{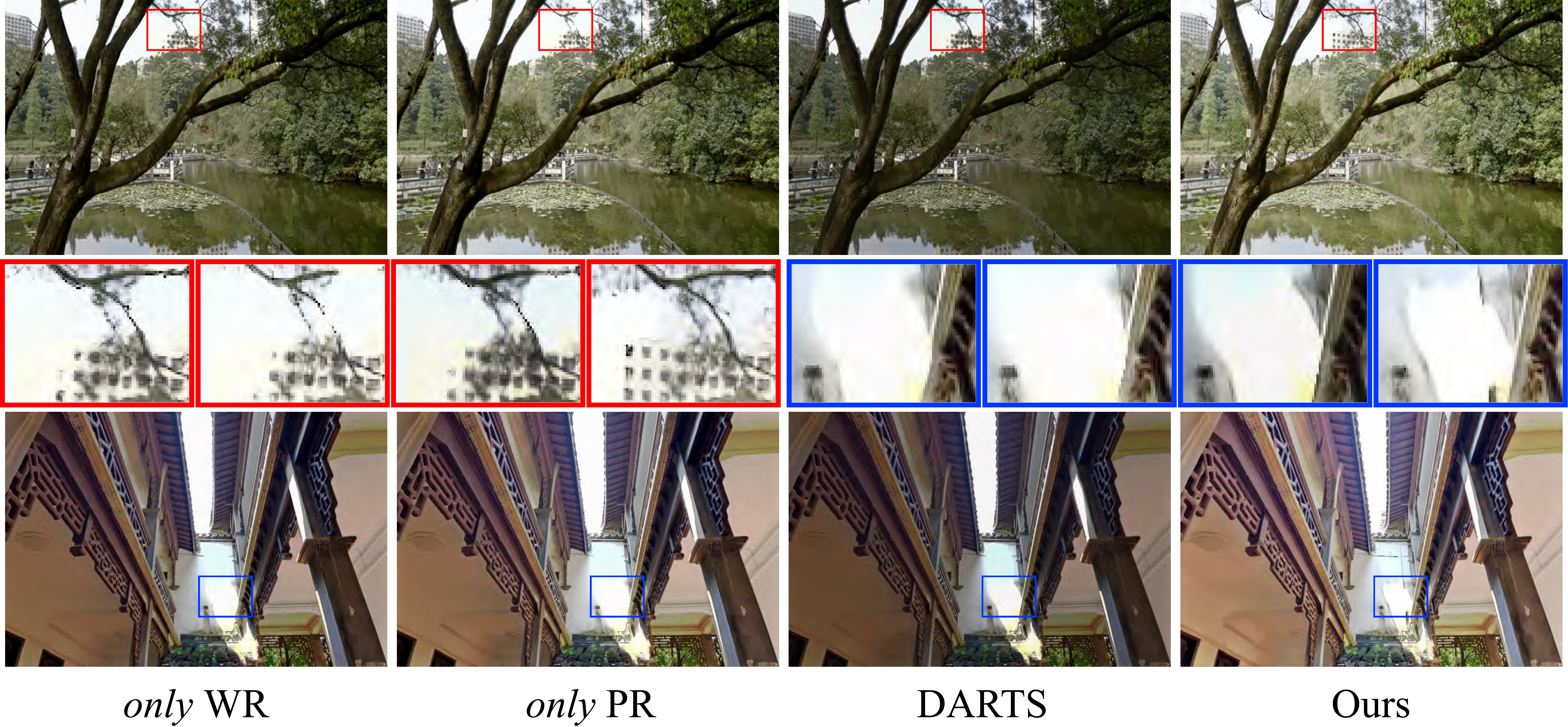} % Reduce the figure size so that it is slightly narrower than the column.
	\caption{Display of visualization results of neural architecture search variants.}
	\label{fig:Ab2}
\end{figure}

We evaluate three other variants: 1) no loss function searching (\emph{w/o} $\boldsymbol{\Gamma}_{\mathtt{H}}$),
2) reference constraint (\emph{only} $\boldsymbol{\Gamma}_{\mathtt{R}}$) and 
3) natural constraint (\emph{only} $\boldsymbol{\Gamma}_{\mathtt{N}}$). 
Qualitative results and quantitative results are shown in Figure.~\ref{fig:Ab1} and Table.~\ref{tab:ab}, respectively. 

Due to the extreme difficulty in designing multiple weights, \emph{w/o} $\boldsymbol{\Gamma}_{\mathtt{H}}$ performs poorly.
The similarity between \emph{only} $\boldsymbol{\Gamma}_{\mathtt{R}}$ and reference image is significantly imporved compared with \emph{only} $\boldsymbol{\Gamma}_{\mathtt{N}}$, since  reference constraint is no longer interfered by  natural constraint, as shown in the error maps.
Meanwhile,  \emph{only} $\boldsymbol{\Gamma}_{\mathtt{R}}$ shows well performance on the SICE dataset with the second highest metrics.
Additionally, \emph{only} $\boldsymbol{\Gamma}_{\mathtt{N}}$ shows better visual effect than other variants on dataset without reference, as shown in the orange circle.
Although the metrics of \emph{only} $\boldsymbol{\Gamma}_{\mathtt{N}}$ is not satisfactory, implicit guidance introduced by natural images enhance the performance of the model, which is reflected in the improvement of the metrics of Ours. In short, the two contrast constraints play a complementary role, and either is indispensable.

\subsubsection{Study on Architecture Searching}
We evaluated three other variants:
1) repealing pruning refinement (\emph{only} WR). 2) repealing weight retention (\emph{only} PR). 3) repealing without both (regular DARTS).
Qualitative results and quantitative results are shown in Figure.~\ref{fig:Ab2} and Table.~\ref{tab:ab}, respectively.

As shown in the red box and blue box, 
since \emph{only} WR cannot eliminate hidden redundant connections between network structures in the search process, while \emph{only} PR can not effectively retain the optimal structure obtained by the search,
their processing of exposure is far worse than Ours, which is reflected in the loss of rich details and the blurring of object boundaries. Their overall visual effect is also poor, and DARTS is even less effective than these two.
Judging from the metrics value, \emph{only} WR is the second highest because weight retention by increasing the number of parameters has an explicit effect, while pruning refine has an implicit effect. In addition, the metrics of \emph{only} PR and \emph{only} WR are both improved compared with DARTS, and are eventually reflected in the improvement of Ours, which prove their effectiveness. In summary, pruning refinement and weight retention are two complementary operations and both have a positive effect on the proposed WSRAS.

\section{Conclusion}
In this paper, we proposed a Hybrid-Supervised Dual-Search approach for MEF, providing a solution for automatically designing loss functions and network structures simultaneously. Through the guidance of hybrid-supervision contrastive constraints, we are able to search for high-precision loss function combinations that break through the upper limit of the reference. With the addition of weighted structure refine architecture search, we discover more efficient and compact network structures. Additionally, we carried out thorough subjective and objective evaluations to underscore the superior efficacy of our approach against various leading-edge methods.

\bibliography{aaai24}

\end{document}